\pdfoutput=1
\documentclass{article}
\usepackage{arxiv}
\usepackage[utf8]{inputenc}
\usepackage[T1]{fontenc}
\usepackage{hyperref}
\usepackage{url}
\usepackage{booktabs}
\usepackage{amsfonts}
\usepackage{makecell}
\usepackage{rotating}
\usepackage{nicefrac}
\usepackage{microtype}
\usepackage{mathtools,bm,amssymb,xcolor}
\usepackage{hyperref}
\usepackage{subcaption,placeins}
\usepackage{adjustbox,multirow,dcolumn}
\usepackage{algorithm,algpseudocode,setspace}
\usepackage{cleveref}
\usepackage{color}
\usepackage{graphicx}
\usepackage[numbers,sort]{natbib}

\usepackage{amsmath,amsfonts,bm}

\def\eqref#1{equation~\ref{#1}}

\def\1{\bm{1}}

\DeclareMathAlphabet{\mathsfit}{\encodingdefault}{\sfdefault}{m}{sl}
\SetMathAlphabet{\mathsfit}{bold}{\encodingdefault}{\sfdefault}{bx}{n}

\graphicspath{{figures/}}

\newcolumntype{d}{D{.}{.}{3.2}}
\makeatletter
\newcolumntype{B}{>{\boldmath\DC@{.}{.}{3.2}}c<{\DC@end}}
\makeatother

\newcommand{\tcenter}[1]{\multicolumn{1}{c}{{#1}}}

\newcommand{\tna}{\tcenter{---}}

\newcommand{\authornote}[1]{\textsuperscript{\rm {#1}}}
\newcommand{\email}[1]{\href{mailto:#1}{\texttt{#1}}}
\newcommand*\samethanks[1][\value{footnote}]{%
    \footnotemark[#1]\hspace{0.4em}}

\makeatletter
\DeclareRobustCommand\onedot{\futurelet\@let@token\@onedot}
\def\@onedot{\ifx\@let@token.\else.\null\fi}
\newcommand{\eg}{\emph{e.g\@\onedot}}

\newcommand{\wrt}{\emph{w.r.t\@\onedot}}

\newcommand{\numero}[1]{No\@.\ {#1}}

\DeclarePairedDelimiter{\parens}{\lparen}{\rparen}

\DeclarePairedDelimiter{\bracks}{[}{]}
\DeclarePairedDelimiter{\braces}{\{}{\}}

\DeclareMathOperator{\loss}{\mathcal{L}}

\makeatletter
\newcommand{\subalign}[1]{%
  \vcenter{%
    \Let@ \restore@math@cr \default@tag
    \baselineskip\fontdimen10 \scriptfont\tw@
    \advance\baselineskip\fontdimen12 \scriptfont\tw@
    \lineskip\thr@@\fontdimen8 \scriptfont\thr@@
    \lineskiplimit\lineskip
    \ialign{\hfil$\m@th\scriptstyle##$&$\m@th\scriptstyle{}##$\hfil\crcr
      #1\crcr
    }%
  }%
}
\makeatother

\newcommand{\inputset}{\mathcal{X}}
\newcommand{\outputset}{\mathcal{Y}}
\newcommand{\cleanset}{\mathcal{D}_\textrm{clean}}
\newcommand{\testset}{\mathcal{D}_\textrm{test}}
\newcommand{\sampledist}{\mathcal{S}}

\newcommand{\ueset}{\mathcal{D}_\textrm{poi}}
\newcommand{\bx}{{\mathbf{x}}}

\newcommand{\btheta}{{\boldsymbol{\theta}}}
\newcommand{\bdelta}{{\boldsymbol{\delta}}}

\newcommand{\expect}{\mathbb{E}}

\title{%
    APBench:
    A Unified Benchmark for
    \mbox{Availability Poisoning Attacks and Defenses}}
\date{}
\author{%
    Tianrui Qin\thanks{%
        Equal contribution.
        Correspondence to Xitong Gao
        (\email{xt.gao@siat.ac.cn}).
    }\hspace{0.4em}\authornote{1,2},
    Xitong Gao\samethanks[1]\authornote{1},
    Juanjuan Zhao\authornote{1},
    Kejiang Ye\authornote{1},
    Cheng-Zhong Xu\authornote{3} \\
    \authornote{1}\,%
        Shenzhen Institutes of Advanced Technology,
        Chinese Academy of Sciences, China. \\
    \authornote{2}\,%
        University of Chinese Academy of Sciences, China. \\
    \authornote{3}\,%
        University of Macau, Macau S.A.R., China.
}

\begin{document}

\maketitle

\begin{abstract}

The efficacy of availability poisoning,
a method of poisoning data by injecting
imperceptible perturbations to prevent its use in model training,
has been a hot subject of investigation.
Previous research suggested that it was difficult
to effectively counteract such poisoning attacks.
However, the introduction of various defense methods
has challenged this notion.
Due to the rapid progress in this field,
the performance of different novel methods
cannot be accurately validated due to variations in experimental setups.
To further evaluate the attack and defense capabilities
of these poisoning methods,
we have developed a benchmark --- APBench
for assessing the efficacy of adversarial poisoning.
APBench consists of 9 state-of-the-art
availability poisoning attacks,
8 defense algorithms,
and 4 conventional data augmentation techniques.
We also have set up experiments
with varying different poisoning ratios,
and evaluated the attacks on multiple datasets
and their transferability across model architectures.
We further conducted a comprehensive evaluation
of 2 additional attacks
specifically targeting unsupervised models.
Our results reveal the glaring inadequacy
of existing attacks
in safeguarding individual privacy.
APBench is open source
and available to the deep learning community:
\url{https://github.com/lafeat/apbench}.

\end{abstract}

\section{Introduction}\label{sec:intro}

Recent advancements
of deep neural networks (DNNs)~\cite{%
    lecun2015deep, schmidhuber2015deep, he2016deep}
heavily rely on the abundant availability
of data resources~\cite{%
    deng2009imagenet,russakovsky2015imagenet,karras2020analyzing}.
However,
the unauthorized collection
of large-scale data through web scraping
for model training
has raised concerns regarding data security and privacy.
In response to these concerns,
a new paradigm of practical
and effective data protection methods
has emerged,
known as availability poisoning attacks
(APA)~\cite{%
    tao2021better,yuan2021neural,fowl2021adversarial,
    huang2021unlearnable,wu2022one,fu2022robust,
    ren2022transferable,he2022indiscriminate,
    sandoval2022autoregressive,feng2019learning,
    yu2022availability, he2022indiscriminate,
    ren2022transferable},
or unlearnable example attacks.
These poisoning methods
inject small
perturbations into images
that are typically imperceptible to humans,
creating ``shortcuts''~\cite{geirhos2020shortcut}
in the training process
that hinder the model's ability
to learn the original features of the images.
Recently,
the field of deep learning
has witnessed advancements
in defense strategies~\cite{%
    liu2023image,qin2023learning,
    dolatabadi2023devil,huang2021unlearnable}
that hold the potential to challenge APAs,
thereby undermining their claimed effectiveness and robustness.
These defenses
reveal the glaring inadequacy of existing APAs
in safeguarding individual privacy in images.
Consequently,
we anticipate an impending arms race
between attack and defense strategies
in the near future.

However,
evaluating the performance
of these new methods
across diverse model architectures and datasets
poses a significant challenge
due to variations in experimental settings
of recent literatures.
In addition,
researchers face the daunting task
of staying abreast of the latest methods
and assessing the effectiveness
of various competing attack-defense combinations.
This could greatly
hamper the development and empirical exploration
of novel attack and defense strategies.

To tackle this challenge,
we propose the APBench,
a benchmark specifically designed
for availability poisoning attacks and defenses.
It involves implementing
poisoning attack and defense mechanisms
under standardized perturbations
and training hyperparameters,
in order to ensure fair
and reproducible comparative evaluations.
APBench comprises a range
of availability poisoning attacks
and defense algorithms,
and commonly-used data augmentation policies.
This comprehensive suite
allows us to evaluate the effectiveness
of the poisoning attacks thoroughly.

Our contributions can be summarized as follows:
\begin{itemize}

    \item An open source benchmark
    for state-of-the-art
    availability poisoning attacks and defenses,
    including 9 supervised and 2 unsupervised
    poisoning attack methods,
    8 defense strategies
    and 4 common data augmentation methods.

    \item
    We conduct a comprehensive evaluation
    of competitions between pairs
    of poisoning attacks and defenses.

    \item
    We conducted experiments
    across 4 publicly available datasets,
    and also extensively examined scenarios
    of partial poisoning,
    increased perturbations,
    the transferability of attacks
    to 4 different DNN models
    under various defenses,
    and unsupervised learning.
    We provide visual evaluation tools
    such as t-SNE, Shapley value map and Grad-CAM
    to qualitatively analyze the impact of poisoning attacks.

\end{itemize}

The aim of APBench
is to serve as a catalyst
for facilitating and promoting future advancements
in both availability poisoning attack and defense methods.
By providing a platform for evaluation and comparison,
we aspire to pave the way for the development
of future availability poisoning attacks
that can effectively preserve utility and protect privacy.

\section{Related Work}\label{sec:related}

\subsection{Availability Poisoning Attacks}

\begin{figure}[ht]
    \centering
    \includegraphics[width=0.95\linewidth]{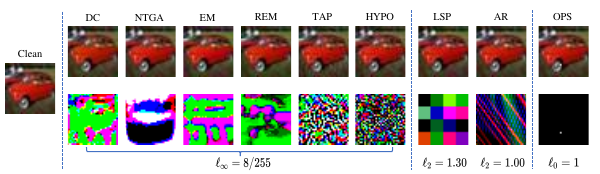}
    \caption{%
        Visualizations of unlearnable CIFAR-10 images
        with corresponding perturbations.
        Perturbations are normalized for visualization.
    }\label{fig:ue_visualize}
\end{figure}

Availability poisoning attacks
belong to a category
of data poisoning attacks~\cite{goldblum2022dataset}
that adds a small perturbation to images,
that is often imperceptible to humans.
The purpose of these perturbations
is to protect individual privacy
from deep learning algorithms,
preventing DNNs from effectively
learning the features present in the images.
To enforce a small perturbation budget,
recent methods typically constrain
their perturbations
within a small \( \ell_p \)-ball
of  \( \epsilon \) radius,
where typically \( p \in \braces{0, 2, \infty} \).
DeepConfuse (DC)~\cite{feng2019learning}
proposes to use autoencoders
to generate training-phase adversarial perturbations.
Neural tangent generalization attacks (NTGA)~\cite{yuan2021neural}
approximates the target model
as a Gaussian process~\cite{jacot2018neural}
using the generalized neural tangent kernel,
and solves a bi-level optimization
for perturbations.
Error-minimizing attacks (EM)~\cite{huang2021unlearnable}
minimizes the training error of the perturbed images
relative to their original labels on the target model,
creating shortcuts for the data
to become ``unlearnable'' by the target model.
Building upon EM,
robust error-minimizing attacks (REM)~\cite{fu2022robust}
use adversarially trained models to generate perturbations
in order to counter defense with adversarial training.
Hypocritical~\cite{tao2021better}
also generates error-minimizing perturbations
similar to EM,
but instead uses a pretrained surrogate model.
Targeted adversarial poisoning (TAP)~\cite{fowl2021adversarial}
uses adversarial examples
as availability attacks.
In contrast to the above approaches,
indiscriminate poisoning (UCL)~\cite{he2022indiscriminate}
and transferable unlearnable examples (TUE)~\cite{ren2022transferable}
instead consider availability poisoning for unsupervised learning.
On the other hand, \(\ell_2\) and \(\ell_0\)
perturbation-based poisoning methods do not require a surrogate model.
They achieve poisoning by searching for certain triggering patterns
to create shortcuts in the network.
Besides the above \( \ell_\infty \)-bounded methods,
Linear-separable poisoning (LSP)~\cite{yu2022availability}
and Autoregressive Poisoning (AR)~\cite{sandoval2022autoregressive}
both prescribe perturbations
within an \(\ell_2\) perturbation budget.
Specifically,
LSP generates randomly initialized linearly
separable color block perturbations,
while AR fills the starting rows and columns
of each channel
with Gaussian noise
and uses an autoregressive process
to fill the remaining pixels,
generating random noise perturbations.
One Pixel Shortcut~\cite{wu2022one} (OPS),
as an \( \ell_0 \)-bounded poisoning method,
perturbs only a single pixel in the training image
to achieve strong poisoning in terms of usability.
\Cref{fig:ue_visualize}
provides visual examples of these attacks.

\subsection{Availability Poisoning Defenses}

Currently, defense methods
against perturbative availability poisoning
can be mainly classified into two categories:
preprocessing and training-phase defenses.
Data preprocessing methods
preprocess the training images
to eliminate the poisoning perturbations
prior to training.
Image shortcuts squeezing (ISS)~\cite{liu2023image}
consists of simple countermeasures
based on image compression,
including grayscale transformation,
JPEG compression,
or bit-depth reduction (BDR)
to perform poison removal.
Recently,
AVATAR~\cite{dolatabadi2023devil}
leverages the method proposed
in DiffPure~\cite{nie2022diffusion}
to employ diffusion models
to disrupt deliberate perturbations
while preserving semantics
in the training images.
On the other hand,
training-phase defense algorithms
apply specific modifications
to the training phase to defense
against availability attacks.
Adversarial training
has long been considered
the most effective defense mechanism~\cite{%
    huang2021unlearnable,fu2022robust}
against such attacks.
Adversarial augmentations~\cite{qin2023learning}
sample multiple augmentations
on one image,
and train models on the maximum loss
of all augmented images
to prevent learning from poisoning shortcuts.
For referential baselines,
APBench also includes
commonly used data augmentation techniques
such as Gaussian blur,
random crop and flip (standard training),
CutOut~\cite{devries2017improved},
CutMix~\cite{yun2019cutmix},
and MixUp~\cite{zhang2017mixup},
and show their (limited) effect
in mitigating availability poisons.

\subsection{Related Benchmarks}

Availability poisoning
is closely connected to the domains
of adversarial and backdoor
attack and defense algorithms.
Adversarial attacks primarily
aim to deceive models with adversarial perturbations
during inference
to induce misclassifications.
Currently,
there are several libraries and benchmarks available
for evaluating adversarial attack and defense techniques,
such as Foolbox~\cite{rauber2020foolbox},
AdvBox~\cite{goodman2020advbox},
and RobustBench~\cite{croce2020robustbench}.

In contrast,
backdoor attacks
focus on injecting backdoor triggers
into the training data,
causing trained models to misclassify images
containing these triggers
while maintaining or minimally impacting clean accuracy.
Benchmark libraries specifically designed
for backdoor attacks and defenses
include TrojanZoo~\cite{pang2020trojanzoo},
Backdoorbench~\cite{wu2022backdoorbench},
and Backdoorbox~\cite{li2023backdoorbox}.

However,
there is currently a lack
and an urgent need
of a dedicated and comprehensive benchmark
that standardizes and evaluates
availability poisoning attack and defense strategies.
To the best of our knowledge,
APBench is the first benchmark
that fulfills this purpose.
It offers an extensive library
of recent attacks and defenses,
explores various perspectives,
including the impact of poisoning rates
and model architectures,
as well as attack transferability.
We hope that APBench
can make significant contributions
to the community
and foster the development
of future availability attacks
for effective privacy protection.

\section{Problem formulation}

\textbf{Attacker}
In general,
the objective of the attacker
is to render the data unlearnable by deep learning models.
They achieve this
by introducing small perturbations to images,
effectively hindering the trainer
from utilizing the data
to learn models that can generalize effectively
to the original data distribution.
Formally,
consider a source dataset
comprising original examples
\(
    \cleanset = \braces*{
        \parens*{\bx_{1}, y_{1}},
        \ldots,\parens*{\bx_{n}, y_{n}}
    }
\)
where \( \bx_i \in \inputset \)
denotes an input image
and \( y_i \in \outputset \) represents its label.
The objective of the attacker
is thus to construct a set
of availability perturbations
\( \bdelta \),
such that models trained on the set
of \emph{availability poisoned examples}
\(
    \ueset\parens{\bdelta} = \braces*{
        \parens{\bx + \bdelta_\bx, y} \mid
        \parens{\bx, y} \in \cleanset
    }
\)
are expected to perform poorly
when evaluated on a test set \( \testset \)
sampled from the distribution \( \sampledist \):
\begin{equation}
    \max_{\bdelta}
    \expect_{
        (\bx_i, y_i) \sim \testset
    }\bracks{
        \loss\parens{
            f_{\btheta^\star\parens{\bdelta}} \parens{\bx_i},
            y_i
        }
    },
    \text{ s.t. }
    \btheta^\star\parens{\bdelta}
        = \underset{\btheta}{\operatorname{argmin}}\,
    \expect_{
        {\parens*{\hat\bx_{i}, y_{i}}
        \sim \ueset \parens{\bdelta}}
    }
    \loss \parens*{
        f_{\btheta} \parens*{ \hat\bx_{i} },
        y_{i}
    },
    \label{eq:attacker}
\end{equation}
where \( \loss \)
denotes the loss function,
usually the softmax cross-entropy loss.
In order to limit the impact
on the original utility of images,
the perturbation \( \bdelta_i \)
is generally constrained
within a small \( \epsilon \)-ball
of \( \ell_p \) distance.

\textbf{Defender}
The objective of the defender
is thus to find a novel training algorithm
\( g(\ueset) \) that trains models
to generalize well
to the original data distribution:
\begin{equation}
    \min_{g} \expect_{
        (\bx_i, y_i) \sim \testset
    }\bracks{
        \loss\parens{
            f_{\btheta^\star} \parens{\bx_i}, y_i
        }
    },
    \text{ s.t. }
    \btheta^\star = g\parens{\ueset}.
    \label{eq:defender}
\end{equation}
Notably,
if the method employs the standard training loss
but performs novel image transformations \( h \),
then \( g \) can be further specialized as follows:
\begin{equation}
    g\parens{\ueset}
    = \underset{\btheta}{\operatorname{argmin}}\,
        \expect_{
            {\parens*{\hat\bx_{i}, y_{i}}
            \sim \ueset \parens{\bdelta}}
        }
        \loss \parens*{
            f_{\btheta} \parens*{
                h\parens*{\hat\bx_{i}}
            }, y_{i}
        }.
    \label{eq:defender:transforms}
\end{equation}

\section{A Unified Availability Poisoning Benchmark}\label{sec:bench}

\begin{figure}[ht]
    \centering
    \includegraphics[
        width=0.90\linewidth, trim=0 10pt 0 10pt
    ]{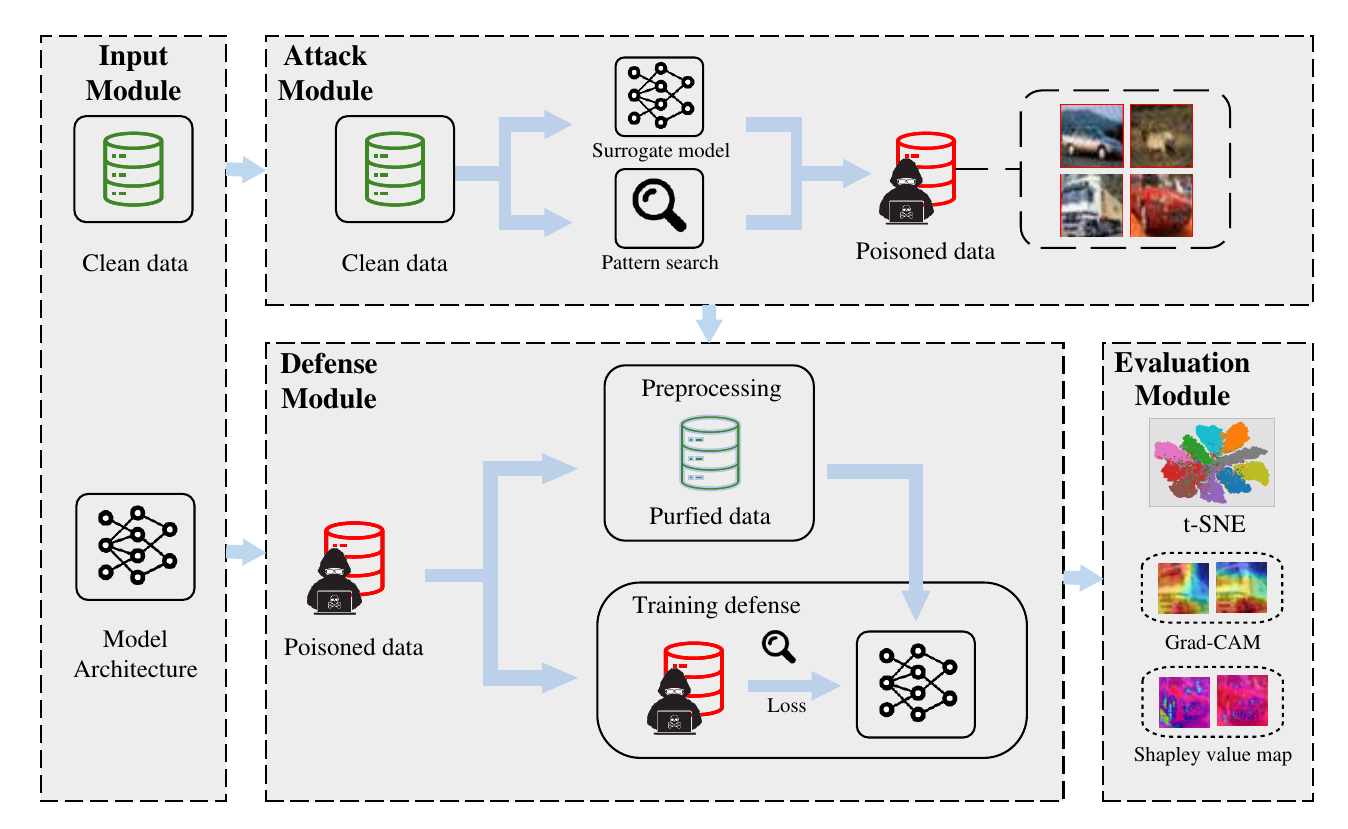}
    \caption{%
        The overall system design of APBench.
    }\label{fig:overview}
\end{figure}

As shown in~\Cref{fig:overview},
APBench consists of three main components:
(a) The availability poisoning attack module.
This library includes a set
of representative availability poisoning attacks
that can generate unlearnable versions
of a given clean dataset.
(b) The poisoning defense module.
This module integrates a suite
of state-of-the-art defenses
that can effectively mitigate the unlearning effect
and restore clean accuracies to a certain extent.
(c) The evaluation module.
This module can efficiently analyze
the performance of various
availability poisoning attack methods
using accuracy metrics
and visual analysis strategies.


We built an extensible codebase
as the foundation of APBench.
In the attack module,
we provide a total of 9 availability poisoning attacks
of 3 different perturbation types (\( \ell_p \))
for supervised learning,
and 2 attacks for unsupervised learning.
For each availability poisoning attack method,
we can generate their respective poisoned datasets.
This module also allows us
to further expand to different perturbations budgets,
poisoning ratios,
and easily extend to future poisoning methods.
Using the poisoned datasets generated
by the attack module,
we can evaluate defenses
through the defense module.
The goal of this module
is to ensure that models trained
on unlearnable datasets
can still generalize well on clean data.
The defense module primarily achieves
poisoning mitigation
through data preprocessing
or training-phase defenses.
Finally,
the evaluation module
computes the accuracy metrics
of different attacks and defense combinations,
and can also perform qualitative visual analyses
to help understand the characteristics
of the datasets.


Our benchmark currently
includes 9 supervised and 2 unsupervised
availability poisoning attacks
and 8 defense algorithms,
and 4 traditional image augmentation methods.
In~\Cref{tab:properties:attacks}
and \Cref{tab:properties:defenses},
we provide a brief summary
of the properties of attack and defense algorithms.
More detailed descriptions for
each algorithm are provided
in \Cref{app:methods}.
\begin{table}[t]
\centering
\caption{%
    Availability poisoning attack algorithms
    implemented in APBench.
    ``Type'' and ``Budget'' respectively
    denotes the type of perturbation and its budget.
    ``Mode'' denotes the training mode,
    where ``S'' and ``U'' and respectively
    mean supervised and unsupervised training.
    ``No surrogate''
    denotes whether the attack requires
    access to a surrogate model for perturbation generation.
    ``Class-wise'' and ``Sample-wise''
    indicate if the attack
    supports class-wise and sample-wise
    perturbation generation.
    ``Stealthy'' denotes
    whether the attack is stealthy to human.
}\label{tab:properties:attacks}
\centering\small
\adjustbox{max width=\linewidth}{%
\begin{tabular}{l|c|c|ccccccc}
\toprule
    Attack Method & Type & Budget & Mode & No surrogate
    & Class-wise & Sample-wise & Stealthy \\
\midrule
DC~\cite{feng2019learning}
    & \multirow{8}{*}{\(\ell_\infty\)} & \multirow{8}{*}{8/255}
    & S & & & \(\checkmark\) & \(\checkmark\) \\
NTGA~\cite{yuan2021neural}
    & & & S & & & \(\checkmark\) & \(\checkmark\) \\
HYPO~\cite{tao2021better}
    & & & S & & & \(\checkmark\) & \(\checkmark\) \\
EM~\cite{huang2021unlearnable}
    & & & S & & \(\checkmark\) & \(\checkmark\) & \(\checkmark\) \\
REM~\cite{fu2022robust}
    & & & S & & \(\checkmark\) & \(\checkmark\) & \(\checkmark\) \\
TAP~\cite{fowl2021adversarial}
    & & & S & & & \(\checkmark\) & \(\checkmark\) \\
UCL~\cite{he2022indiscriminate}
    & & & U & & \(\checkmark\) & \(\checkmark\) & \(\checkmark\) \\
TUE~\cite{ren2022transferable}
    & & & U & & &\(\checkmark\) & \(\checkmark\) \\
\midrule
LSP~\cite{yu2022availability}
    & \multirow{2}{*}{\(\ell_2\)} & 1.30
    & S & \(\checkmark\) & \(\odot\) & \(\checkmark\) & \\
AR~\cite{sandoval2022autoregressive}
    & & 1.00
    & S & \(\checkmark\) & \(\odot\) & \(\checkmark\) & \(\checkmark\) \\
\midrule
OPS~\cite{wu2022one}
    & \(\ell_0\) & 1 & S & \(\checkmark\) & \(\checkmark\) & \(\checkmark\) & \\
\bottomrule
\end{tabular}}
\centering\small
\caption{%
   Availability poisoning defense algorithms
   implemented in APBench.
}\label{tab:properties:defenses}
\adjustbox{max width=\linewidth}{%
\begin{tabular}{l|l|l|l}
\toprule
Defense Method & Type & Time Cost & Description  \\
\midrule
Standard
    & \multirow{4}{*}{Data augmentations}
    & Low & Random image cropping and flipping \\
CutOut~\cite{devries2017improved}
    & & Low & Random image erasing \\
MixUp~\cite{zhang2017mixup}
    & & Low & Random image blending \\
CutMix~\cite{yun2019cutmix}
    & & Low & Random image cutting and stitching \\
\midrule
Gaussian (used in~\cite{liu2023image})
    & \multirow{5}{*}{Data preprocessing}
    & Low & Image blurring with a Gaussian kernel \\
BDR (used in~\cite{liu2023image})
    & & Low & Bit-depth reduction \\
Gray~(used in \cite{liu2023image})
    & & Low & Image grayscale transformation \\
JPEG~(used in \cite{liu2023image})
    & & Low & Image compression \\
AVATAR~\cite{dolatabadi2023devil}
    & & High & Image corruption and restoration \\
\midrule
U-Lite~\cite{qin2023learning}
    & \multirow{3}{*}{Training-phase defense}
    & Low & Stronger data augmentations \\
U-Max~\cite{qin2023learning}
    & & High & Adversarial augmentations \\
AT~\cite{madry2017towards}
    & & High & Adversarial training \\
\bottomrule
\end{tabular}}
\end{table}

\section{Evaluations}\label{sec:eval}


\textbf{Datasets}
We evaluated our benchmark
on 4 commonly used datasets
(CIFAR-10~\cite{krizhevsky2009learning},
 CIFAR-100~\cite{krizhevsky2009learning},
 SVHN~\cite{netzer2011reading},
 and an ImageNet~\cite{deng2009imagenet} subset)
and 5 mainstream models
(ResNet-18~\cite{he2016deep}, ResNet-50~\cite{he2016deep},
 MobileNetV2~\cite{sandler2018mobilenetv2},
 and DenseNet-121~\cite{huang2017densely}).
To ensure a fair comparison
between attack and defense methods,
we used only the basic version of training
for each model.
\Cref{app:datasets}
summarizes the specifications of the datasets
and the test accuracies
achievable through standard training
on clean training data,
and further describes the detail specifications
of each dataset.

\textbf{Attacks and defenses}
We evaluated combinations of availability poisoning attacks
and defense methods introduced in \Cref{sec:bench}.
Moreover, we explored 5 different data poisoning rates
and 5 different models.
In addition, We also explore
two availability poisonings
for unsupervised learning
(UCL~\cite{he2022indiscriminate} and TUE~\cite{ren2022transferable})
and evaluate them
on the recently proposed defenses
(Gray, JPEG, UEraser-Lite~\cite{qin2023learning},
 and AVATAR~\cite{dolatabadi2023devil}).
The implementation details
of all algorithms and additional results
can be found in \Cref{app:methods}.

\textbf{Training settings}
We trained the CIAFR-10,
CIFAR-100 and ImageNet-subset models for 200 epochs
and the SVHN models for 100 epochs.
We used the stochastic gradient descent (SGD) optimizer
with a momentum of 0.9
and a learning rate of 0.1
by default.
As for unsupervised learning,
all experiments are trained for 500 epochs
with the SGD optimizer.
The learning rate is 0.5
for SimCLR~\cite{chen2020simple}
and 0.3 for MoCo-v2~\cite{chen2020improved}.
Please note that we generate sample-wise perturbations
for all availability poisoning attacks.
Specific settings for each defense method
may have slight differences,
and detailed information
can be found in the~\Cref{app:settings}.

\subsection{Standard Scenario}
To start,
we consider a common scenario
where both the surrogate model
and target model are ResNet-18,
and the poisoning rate is set to \(100\%\).
We first evaluate the performance
of the supervised poisoning methods
against 4 state-of-the-art defense mechanisms
and 4 commonly used data augmentation strategies.
\Cref{tab:c10} presents the evaluation results
on CIFAR-10 from our benchmark.
It is evident that the conventional data augmentation methods
appear to be ineffective
against all poisoning methods.
Yet, even simple image compression methods
(BDR, grayscale, and JPEG corruption from ISS~\cite{liu2023image})
demonstrate a notable effect
in mitigating the poisoning attacks,
but fails to achieve high clean accuracy.
Despite requiring more computational cost
or additional resources
(pretrained diffusion models for AVATAR),
methods such as UEraser-Max~\cite{qin2023learning}
and AVATAR~\cite{dolatabadi2023devil},
generally surpass the image compression methods from ISS
in terms of effectiveness.
Adversarial training appears effective
but in many cases is outperformed
by even a simple JPEG compression.
We further conduct experiments
on the CIFAR-100, SVHN, and ImageNet-subset datasets,
and the results are shown in~\Cref{tab:adddataset}.
\begin{table}[t]
\vspace{-12pt}
\centering
\caption{%
    Test accuracies (\%)
    of models trained
    on poisoned CIFAR-10 datasets.
}\label{tab:c10}
\adjustbox{max width=\linewidth}{%
\begin{tabular}{l|ccccc|cccccc}
\toprule
{Method}
    & Standard & CutOut & CutMix & MixUp & Gaussian
    & BDR & Gray & JPEG & U-Max & AVATAR & AT \\
\midrule
DC
    & 15.19 & 19.94 & 17.91 & 25.07 & 16.10
    & 67.73 & 85.55 & 83.57 & 92.17 & 82.10 & 76.85 \\
EM
    & 20.78 & 18.79 & 22.28 & 31.14 & 14.71
    & 37.94 & 92.03 & 80.72 & 93.61 & 75.62 & 82.51 \\
REM
    & 17.47 & 21.96 & 26.22 & 43.07 & 21.80
    & 58.60 & 92.27 & 85.44 & 92.43 & 82.42 & 77.46 \\
HYPO
    & 70.38 & 69.04 & 67.12 & 74.25 & 62.17
    & 74.82 & 63.35 & 85.21 & 88.44 & 85.94 & 81.49 \\
NTGA
    & 22.76 & 13.78 & 12.91 & 20.59 & 19.95
    & 59.32 & 70.41 & 68.72 & 86.78 & 86.22 & 69.70 \\
TAP
    & 6.27 & 9.88 & 14.21 & 15.46 & 7.88
    & 70.75 & 11.01 & 84.08 & 79.05 & 87.75 & 79.92 \\
\midrule
LSP
    & 13.06 & 14.96 & 17.69 & 18.77 & 18.61
    & 53.86 & 64.70 & 80.14 & 92.83 & 76.90 & 81.38 \\
AR
    & 11.74 & 10.95 & 12.60 & 14.15 & 13.83
    & 36.14 & 35.17 & 84.75 & 90.12 & 88.60 & 81.15 \\
\midrule
OPS
    & 14.69 & 52.98 & 64.72 & 49.27 & 13.38
    & 37.32 & 19.88 & 78.48 & 77.99 & 66.16 & 14.95 \\
\bottomrule
\end{tabular}}
\centering
\caption{%
    Test accuracies (\%)
    on poisoned CIFAR-100, SVHN and ImageNet-subset datasets.
}\label{tab:adddataset}
\adjustbox{max width=\linewidth}{%
\begin{tabular}{ll|ccccc|cccc}
\toprule
    Dataset & {Method}
    & Standard & CutOut & CutMix & MixUp & Gaussian
    & BDR & Gray & JPEG & U-max \\
\midrule
\multirow{4}{*}{CIFAR-100}
& EM
    & 3.03 & 4.15 & 3.98 & 6.46 & 2.99
    & 34.10 & 59.14 & 58.71 & 68.81 \\
& REM
    & 3.73 & 4.00 & 3.71 & 10.90 & 3.59
    & 29.16 & 57.47 & 55.60 & 67.72 \\
& LSP
    & 2.56 & 2.33 & 4.52 & 4.86 & 1.71
    & 27.12 & 39.45 & 52.82 & 68.31 \\
& AR
    & 1.87 & 1.63 & 3.17 & 2.35 & 2.62
    & 31.15 & 16.13 & 54.73 & 55.95 \\
\midrule
\multirow{4}{*}{SVHN}
& EM
    & 10.33 & 13.38 & 10.77 & 12.79 & 8.82
    & 36.65 & 65.66 & 86.14 & 90.24 \\
& REM
    & 14.02 & 18.92 & 9.55 & 19.56 & 7.54
    & 42.52 & 19.59 & 90.58 & 88.26 \\
& LSP
    & 12.16 & 12.98 & 8.17 & 18.86 & 7.15
    & 26.67 & 16.90 & 84.06 & 90.64 \\
& AR
    & 19.23 & 14.92 & 6.71 & 13.52 & 7.75
    & 39.24 & 10.00 & 92.46 & 90.07 \\
\midrule
\multirow{3}{*}{ImageNet-100}
& EM
    & 2.94 & 4.05 & 4.73 & 4.15 & 3.15
    & 6.45 & 12.20 & 31.73 & 44.07 \\
& REM
    & 3.66 & 4.13 & 4.78 & 3.94 & 4.28
    & 4.03 & 3.95 & 40.98 & 42.14 \\
& LSP
    & 38.52 & 40.56 & 29.78 & 7.85 & 42.68
    & 26.58 & 25.18 & 36.83 & 63.28 \\
\bottomrule
\end{tabular}}
\end{table}

\subsection{Challenging Scenarios}\label{sec:eval:challenging}

To further investigate the effectiveness
and robustness of availability poisoning attacks and defenses,
we conducted evaluations in more challenging scenarios.
We considered partial poisoning scenarios,
larger perturbation poisoning,
and the attack transferability to different models.

\begin{figure*}[htbp]
\centering
\includegraphics[width=0.7\linewidth]{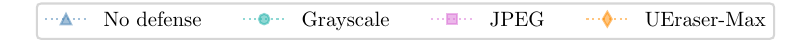}
\newcommand{\ppsubfig}[2]{%
    \begin{subfigure}{0.25\linewidth}
        \includegraphics[
            width=\linewidth, trim=10pt 10pt 5pt 10pt, clip
        ]{fig/partial/#1.pdf}
        \caption{#2.}\label{fig:partial:#1}
    \end{subfigure}%
}
\ppsubfig{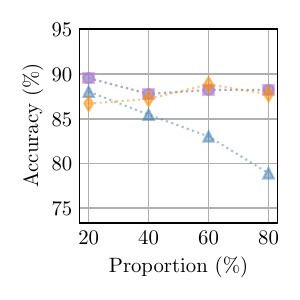}{EM}%
\ppsubfig{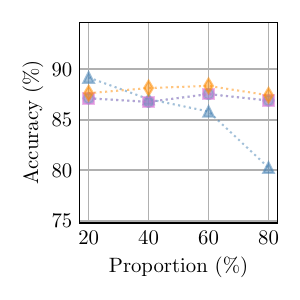}{REM}%
\ppsubfig{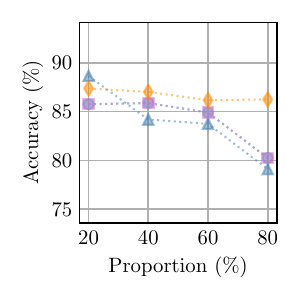}{LSP}%
\ppsubfig{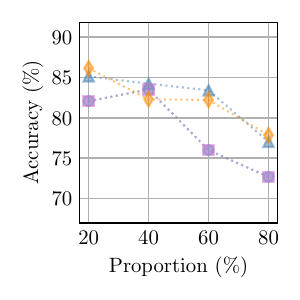}{AR}%
\caption{%
    The efficacy of defenses
    (No defense, Grayscale, JPEG, and UEraser-Max)
    against different partial poisoning attacks
    including EM (\subref{fig:partial:em}),
    REM (\subref{fig:partial:rem}),
    LSP (\subref{fig:partial:lsp}),
    and AR (\subref{fig:partial:ar})
    with poisoning ratios ranging from 20\% to 80\%.
}\label{fig:partial}
\end{figure*}

\textbf{Partial poisoning}
Here, we investigate the impact of poisoning rate
on the performance of availability poisoning.
\Cref{fig:partial}
presents the results on CIFAR-10 and ResNet-18,
\wrt{} each poisoning rate for attack-defense pairs,
where each subplot corresponds to a specific poisoning attack method.
We explore four different poisoning rates
(20\%, 40\%, 60\%, 80\%).
It is evident
that due to the different objectives
of availability poisoning
compared to traditional data poisoning,
partial poisoning can greatly affect
the effectiveness of availability poisoning attacks.
A small poisoning rate
mostly nullified their effectiveness.

\textbf{Larger perturbations}
We increased the magnitude of perturbations
in availability poisoning attacks
to further evaluate the performance of attacks and defenses.
\Cref{tab:large_pert} presents the results of
availability poisoning with larger perturbations on CIFAR-10.
Due to such significant perturbations,
their stealthiness is further reduced,
making it challenging to carry out such attacks
in realistic scenarios.
However, larger perturbations indeed
have a more pronounced impact on suppressing defense performance,
leading to significant accuracy losses for all defense methods.
There exists a trade-off between perturbation magnitude and accuracy recovery.
\begin{table}[t]
\centering\small
\caption{%
    Test accuracies (\%)
    on poisoned CIFAR-10 datasets with increased perturbations.
}\label{tab:large_pert}
\begin{tabular}{l|lccccc}
\toprule
 {Method} & Budget
 & No defense & Gray & JPEG & U-max & AT\\
\midrule
EM & {\(\ell_\infty=16/255\)} & 18.74 & 76.76 & 55.96 & 88.09 & 77.82 \\
REM & {\(\ell_\infty=16/255\)} & 19.80 & 83.65 & 80.07 & 80.36 & 75.64 \\
\midrule
LSP & \(\ell_2=1.74\) & 15.83 & 37.60 & 42.83 & 87.20 & 77.92\\
AR & \(\ell_2=1.50\) & 11.20 & 26.10 & 78.24 & 68.42 & 70.14\\
\bottomrule
\end{tabular}
\end{table}

\textbf{Attack transferability across models}
In real-world scenarios,
availability poisoning attackers
can only manipulate the data and do not have access
to specific details of the defender.
Therefore, we conducted experiments on different model architectures.
It is worth noting that
all surrogate-based attack methods
are considered using ResNet-18.
The results are shown in~\Cref{tab:arch}.
It is evident that all surrogate-based
and -free poisoning methods
exhibit strong transferability,
while the three recently proposed defenses
also achieve successful defense
across different model architectures.
\begin{table}[t]
\vspace{-8pt}
\centering\small
\caption{%
     Clean test accuracy of CIFAR-10 target models in the transfer setting.
     Note that AR and LSP are surrogate-free
     and the surrogate model is ResNet-18 for EM and REM.
}\label{tab:arch}
\begin{tabular}{ll|ccccc}
\toprule
Method
 & Model & No defense & Gray & JPEG & U-max & AVATAR \\
\midrule
\multirow{4}{*}{EM}
 & ResNet-50 & 14.41 & 83.40 & 76.88 & 85.89 & 77.64 \\
 & SENet-18 & 13.60 & 86.03 & 79.35 & 83.27 & 74.22 \\
 & MobileNetV2 & 15.62 & 77.21 & 70.96 & 82.71 & 75.62 \\
 & DenseNet-121 & 13.89 & 82.49 & 78.42 & 82.37 & 76.69 \\
 \midrule
\multirow{4}{*}{REM}
 & ResNet-50 & 16.26 & 87.26 & 75.79 & 92.69 & 83.68\\
 & SENet-18 & 20.99 & 84.50 & 78.92 & 93.17 & 84.37\\
 & MobileNetV2 & 20.83 & 80.81 & 72.27 & 91.03 & 82.77\\
 & DenseNet-121 & 21.45 & 85.47 & 78.42 & 93.09 & 83.04\\
\midrule
\multirow{4}{*}{LSP}
 & ResNet-50 & 19.23 & 68.94 & 73.24 & 93.08 & 76.47\\
 & SENet-18 & 18.54 & 65.06 & 76.51 & 92.53 & 75.19\\
 & MobileNetV2 & 16.82 & 61.07 & 72.03 & 92.10 & 76.81 \\
 & DenseNet-121 & 18.94 & 67.95 & 74.90 & 93.47 & 78.22 \\
\midrule
\multirow{4}{*}{AR}
 & ResNet-50 & 11.83 & 27.51 & 80.24 & 81.40 & 86.39\\
 & SENet-18 & 13.68 & 34.26 & 79.29 & 75.06 & 84.37 \\
 & MobileNetV2 & 13.36 & 28.54 & 68.14 & 73.40 & 81.63\\
 & DenseNet-121 & 13.43 & 25.51 & 81.12 & 82.36 & 89.92\\
\bottomrule
\end{tabular}
\end{table}

\textbf{Unsupervised learning}
We evaluated the availability poisoning attacks
targeting unsupervised models
on two benchmark datasets (CIFAR-10 and CIFAR-100).
We assumed the victim model to be ResNet-18,
while the attacker used ResNet-18 to generate adversarial perturbations.
We considered two popular unsupervised learning frameworks:
SimCLR~\cite{chen2020simple} and MoCo-v2~\cite{chen2020improved}.
All defense methods were applied before the data augmentation process,
which means they were applied to preprocessed images
before undergoing different data augmentations.
Therefore, we only applied UEraser-lite
as a data preprocessing method.
The results of all experiments are shown in~\Cref{tab:unsuper}.
\begin{table}[t]
\centering\small\caption{%
     Performance of availability poisoning attacks and defense
     on different unsupervised learning algorithms and datasets.
     Note that ``U-lite'' denotes UEraser-lite.
}\label{tab:unsuper}
\begin{tabular}{cc|c|c|c|c|c}
\toprule
    Algorithm & Method & No Defense & Gray & JPEG & U-lite & AVATAR \\
\midrule
    \multirow{2}{*}{SimCLR} & UCL & 47.25 & 46.91 & 66.76 & 68.42 & 83.22\\
 & TUE & 57.10 & 56.37 & 67.54 & 66.59 &84.24 \\
\midrule
    \multirow{2}{*}{MoCo-v2} & UCL & 53.78 & 53.34 & 65.44 & 72.13 &83.08 \\
 & TUE & 66.73 & 64.95 & 67.28 & 74.82 &82.48 \\
\bottomrule
\end{tabular}
\end{table}

\textbf{Visual Analyses}
We provide visualization tools
to facilitate the analysis and understanding
of availability poisoning attacks.
We use t-SNE~\cite{van2008visualizing}
to visualize the availability poisons.
Although t-SNE cannot accurately
represent high-dimensional spaces,
it aids in the global visualization of feature representations,
allowing us to observe specific characteristics
of availability poisons.
Gradient-weighted class activation mapping
(Grad-CAM)~\cite{selvaraju2017grad}
and Shapley value map~\cite{lundberg2017unified}
are commonly used image analysis tools that visualize
the contributions of different pixels in an image
to the model's predictions.
Through Grad-CAM or Shapley value map,
we can qualitatively
compare the performance of test samples
between the poisoned and clean models.
\begin{figure}[!ht]
    \centering%
    \includegraphics[width=\linewidth]{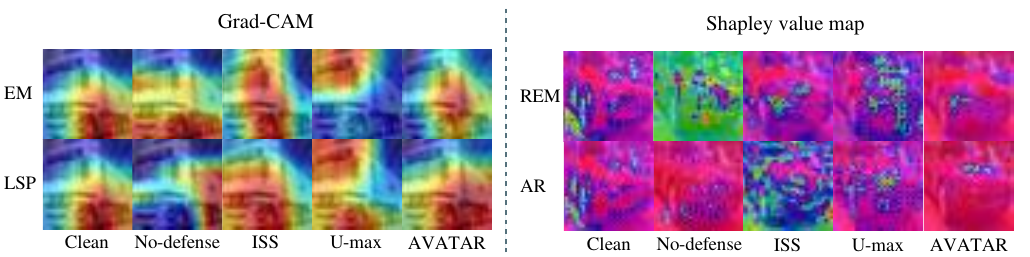}
    \caption{%
        Grad-CAM and Shapley value map visualizations
        of regions contributed to model decision under different attack
        methods and defense methods with ResNet-18.
        (Left)
        The Grad-CAM visualizations of EM and LSP
        poisoning attacks.
        (Right)
        The Shapley value map visualizations of REM and AR
        poisoning attacks.
    }\label{fig:gsimg}
\end{figure}
\begin{figure}[!ht]
    \centering%
    \includegraphics[width=0.90\linewidth]{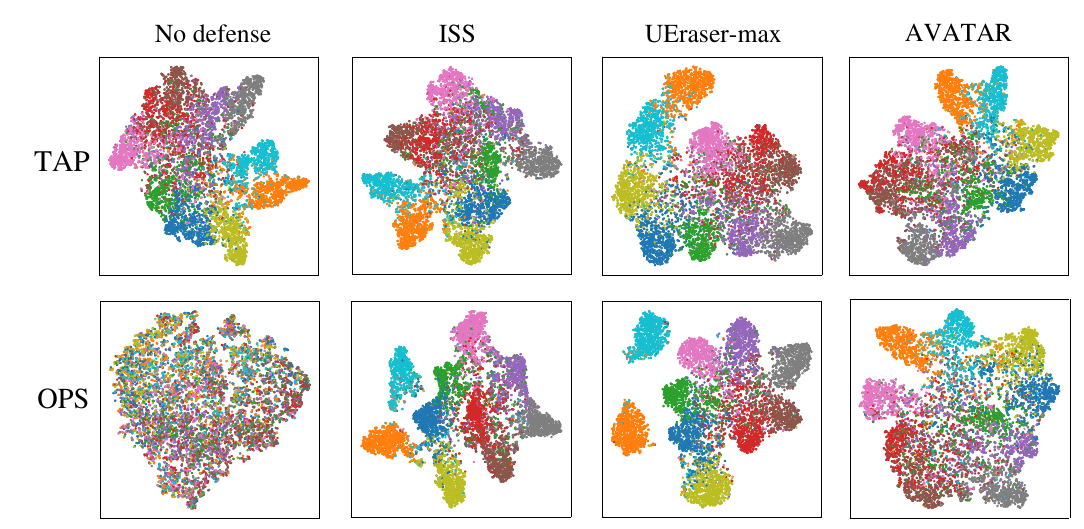}
    \caption{%
        The t-SNE visualization of the models'
        feature representations on the clean test set.
        Note that without defenses,
        the feature representations of the poisoned models
        are mostly scrambled as the models struggle
        to learn useful features.
    }\label{fig:tsne}
\end{figure}

\section{Discussion}

Our findings indicate
that perturbations constrained
by traditional \( \ell_p \) norms
are ineffective
against adversarial augmentation (UEraser),
and image restoration
by pre-trained diffusion models (AVATAR),
as they break free from the assumption
of \( \ell_p \) constraints.
Even simple image compression techniques
(JPEG, Grayscale, and BDR)
can effectively remove the effect of perturbations.
At this stage,
availability poisoning attacks
that rely on \( \ell_p \)-bounded perturbations
may not be as effective as initially suggested
by the relevant attacks.

Future research directions
should explore methods that enhance the resilience of perturbations.
One approach to consider
is the development of generalizable attacks,
which can simultaneously target the DNNs being trained,
diffusion models for image restoration,
and remain robust against traditional or color distortions,
among others.
On the other hand,
semantic-based perturbations
offer an alternative strategy,
as such modifications to images
can be challenging to remove by defenses.

\section{Conclusions}

We have established
the first comprehensive and up-to-date benchmark
for the field of availability poisoning,
covering a diverse range
of availability poisoning attacks
and state-of-the-art defense algorithms.
We have conducted effective evaluations and analyses
of different combinations of attacks and defenses,
as well as additional challenging scenarios.
Through this new benchmark,
our primary objective
is to provide researchers
with a clearer understanding
of the current progress
in the field of availability poisoning attacks and defenses.
We hope it can enable rapid comparisons
between existing methods and new approaches,
while also inspiring fresh ideas
through our comprehensive evaluations and observations.
We believe that our benchmark
will contribute to the advancement
of availability poisoning research
and the development of more effective attacks
to safeguard privacy.

\section*{Acknowledgements}\label{sec:acknowledgements}
This work is supported in part by
Guangdong Basic and Applied Basic Research Foundation
(\numero{2020B1515130004}),
and Shenzhen Science and Technology Innovation Commission
(\numero{JCYJ20190812160003719}).
\bibliographystyle{plain}
\bibliography{references}
\newpage
\appendix
\section{Datasets}\label{app:datasets}

\Cref{tab:datasets}
summarizes the specifications of datasets
and the respective test accuracies
of typical training on ResNet-18 architectures.
\begin{table}[ht]
\centering\caption{%
    Dataset specifications
    and the respective test accuracies on ResNet-18.
}\label{tab:datasets}
\adjustbox{scale=0.8}{%
\begin{tabular}{l|cccc}
\toprule
Datasets
    & \#Classes & Training / Test Size & Image Dimensions & Clean Accuracy (\%) \\
\midrule
CIFAR-10~\cite{krizhevsky2009learning}
    & 10 & 50,000 / 10,000 & 32\(\times\)32\(\times\)3 & 94.32 \\
CIFAR-100~\cite{krizhevsky2009learning}
    & 100 & 50,000 / 10,000 & 32\(\times\)32\(\times\)3 & 75.36 \\
SVHN~\cite{netzer2011reading}
    & 10 & 73,257 / 26,032 & 32\(\times\)32\(\times\)3 & 96.03 \\
ImageNet-subset~\cite{deng2009imagenet}
    & 100 & 20,000 / 4,000 & 224\(\times\)224\(\times\)3 & 72.44 \\
\bottomrule
\end{tabular}}
\end{table}

\section{Implementation Details}\label{app:methods}

In addition to the discussion of properties
of the availability poisoning attacks and defenses
presented in~\Cref{%
    tab:properties:attacks,tab:properties:defenses},
here,
we provide a high-level description
of the attack and defense algorithms
implemented in APBench.

\textbf{Attacks}:
\begin{itemize}

    \item \textbf{Deep Confuse (DC)}~\cite{feng2019learning}:
    DC is proposed as a novel approach
    to manipulating classifiers by modifying the training data.
    Its key idea involves employing
    an autoencoder-like network
    to capture the training trajectory
    of the target model and adversarially perturbing
    the training data.

    \item \textbf{Error-minimizing attack (EM)}~\cite{huang2021unlearnable}:
    EM trains a surrogate model
    by minimizing the error of images
    relative to their original labels,
    generating perturbations
    that minimize the errors
    and thus render the perturbed images unlearnable.
    The authors of EM introduce
    the threat model of availability poisoning attacks,
    highlighting their role
    as a mechanism for privacy protection.

    \item \textbf{%
        Neural tangent generalization attack (NTGA)%
    }~\cite{yuan2021neural}:
    NTGA simulates the training dynamics
    of a generalized deep neural network
    using a Gaussian process
    and leverages this surrogate
    to find better local optima
    with improved transferability.

    \item \textbf{Hypocritical (HYPO)}~\cite{tao2021better}:
    HYPO, similar to EM,
    generates images that minimize errors
    relative to their true labels using a pre-trained model.

    \item \textbf{Targeted adversarial poisoning (TAP)}~\cite{fowl2021adversarial}:
    TAP achieves availability poisoning
    by generating targeted adversarial examples
    of non-ground-truth labels of pre-trained models.

    \item \textbf{Robust error-minimizing attacks (REM)}~\cite{fu2022robust}:
    REM improves the poisoning effect
    of availability poisoning
    by replacing the training process of the surrogate model
    with adversarial training.

    \item \textbf{Linear-separable poisoning (LSP)}~\cite{yu2022availability}:
    LSP generates randomly initialized
    linearly separable color block perturbations,
    enabling effective availability poisoning attacks
    without requiring surrogate models
    or excessive computational overhead.

    \item \textbf{Autoregressive Poisoning (AR)}~\cite{sandoval2022autoregressive}:
    AR, similar to LSP,
    does not require additional surrogate models.
    It fills the initial rows and columns
    of each channel with Gaussian noise
    and uses an autoregressive process
    to fill the remaining pixels,
    generating random noise perturbations.

    \item \textbf{One-Pixel-Shortcut (OPS)}~\cite{wu2022one}:
    OPS is a targeted availability poisoning attack
    that perturbs only one pixel of an image,
    generating an effective availability poisoning attack
    against traditional adversarial training methods.

    \item \textbf{Indiscriminate poisoning (UCL)}~\cite{he2022indiscriminate}:
    UCL considers generating unlearnable examples
    for unsupervised learning
    by minimizing the CL loss (\eg{}, the InfoNCE loss)
    in the unsupervised learning setting.

    \item \textbf{Transferable unlearnable examples (TUE)}~\cite{ren2022transferable}:
    TUE discovers that UCL is effective
    only in unsupervised learning,
    while its performance significantly deteriorates
    in supervised learning.
    Therefore,
    TUE is proposed that simultaneously
    targets both supervised and unsupervised learning.
    Different to UCL,
    it additionally embeds
    linear separable poisons
    into unsupervised unlearnable examples
    using the class-wise separability discriminant.

\end{itemize}

\textbf{Defenses}:
\begin{itemize}

    \item \textbf{Adversarial training (AT)}~\cite{madry2017towards}:
    AT is a widely-recognized effective approach
    against availability poisoning.
    Small adversarial perturbations
    are applied to the training images
    during training,
    in order to improve the robustness
    of the model against perturbations.

    \item \textbf{Image Shortcut Squeezing (ISS)}~\cite{liu2023image}:
    ISS uses traditional image compression techniques
    such as grayscale transformation,
    bit-depth reduction (BDR),
    and JPEG compression,
    as defenses against availability poisoning.

    \item \textbf{Adversarial augmentations (UEraser)}~\cite{qin2023learning}:
    UEraser-Lite uses an effective augmentation pipeline
    to suppress availability poisoning shortcuts.
    UEraser-Max further improves the defense
    against availability poisoning
    through adversarial augmentations.

    \item \textbf{AVATAR}~\cite{dolatabadi2023devil}:
    Following DiffPure~\cite{nie2022diffusion},
    AVATAR cleans the images
    of the unlearnable perturbations
    with diffusion models.

\end{itemize}

\section{Experimental Settings}\label{app:settings}

\Cref{tab:config:attack}
presents the default hyperparameters
for all availability poisoning attacks
implemented in APBench.
\begin{table}[!ht]
\centering\caption{%
    Default hyperparameter settings
    of attack methods.
}\label{tab:config:attack}
\adjustbox{scale=0.8}{%
\begin{tabular}{lll}
\toprule
Methods & Hyperparameter & Settings \\
\midrule
\multirow{2}{*}{DC}
    & Perturbation & \(\ell_\infty = 8/255\) \\
    & Pre-trained model
        & \href{https://github.com/kingfengji/DeepConfuse/blob/master/CIFAR/pretrained/atk.0.032.best.pth}{Official pretrained} \\
\midrule
\multirow{2}{*}{NTGA}
    & Perturbation  & \(\ell_\infty = 8/255\) \\
    & Poisoned dataset
        & \href{https://github.com/lionelmessi6410/ntga/tree/main#unlearnable-datasets}{Official pretrained CIFAR-10 CNN (best)} \\
\midrule
\multirow{6}{*}{EM}
    & Perturbation
    & \(\ell_\infty = 8/255\) \\
    & Perturbation type & Sample-wise \\
    & Stopping error rate & 0.01 \\
    & Learning rate & 0.1 \\
    & Batch size & 128 \\
    & Optimizer & SGD \\
\midrule
\multirow{2}{*}{HYPO}
    & Perturbation & \(\ell_\infty = 8/255\) \\
    & Step size & \(\ell_\infty = 0.8/255\) \\
\midrule
\multirow{1}{*}{TAP}
    & Perturbation & \(\ell_\infty = 8/255\) \\
\midrule
\multirow{7}{*}{REM}
    & Perturbation & \(\ell_\infty = 8/255\) \\
    & Perturbation type & Sample-wise \\
    & Stopping error rate & 0.01 \\
    & Learning rate & 0.1 \\
    & Batch size & 128 \\
    & Optimizer & SGD \\
    & Adversarial training perturbation
        & \(\ell_\infty = 4/255\) \\
\midrule
\multirow{2}{*}{LSP}
    & Perturbation & \(\ell_2 = 1.30\) (Project from \(\ell_\infty = 6/255\)) \\
    & Patch size & 8 for CIFAR-10/100 and SVHN; 32 for ImageNet\\
\midrule
\multirow{2}{*}{AR}
    & Perturbation & \(\ell_2 = 1.00\) \\
    & Default hyperparameters
        & \href{https://github.com/psandovalsegura/autoregressive-poisoning}{Follows official code} \\
\midrule
\multirow{3}{*}{OPS}
    & Perturbation & \(\ell_0 = 1\) \\
    & Perturbation type & Sample-wise \\
    & Default hyperparameters
        & \href{https://github.com/cychomatica/One-Pixel-Shotcut}{Follows official code} \\
\midrule
\multirow{2}{*}{UCL}
    & Perturbation & \(\ell_\infty = 8/255\) \\
    & Poisoned dataset
        & \href{https://github.com/kaiwenzha/contrastive-poisoning}{Official pretrained CP-S of UCL} \\
\midrule
\multirow{2}{*}{TUE}
    & Perturbation & \(\ell_\infty = 8/255\) \\
    & Poisoned dataset
        & \href{https://github.com/renjie3/TUE}{Official pretrained} \\
\bottomrule
\end{tabular}}
\end{table}
\begin{table}[ht]
\centering\caption{%
    Default training hyperparameter settings.
}\label{tab:settings:general}
\adjustbox{scale=0.8}{%
\begin{tabular}{lll}
\toprule
Datasets & Hyperparameter & Settings \\
\midrule
\multirow{8}{*}{CIFAR-10/-100}
    & Optimizer & SGD \\
    & Momentum & 0.9 \\
    & Weight-decay & 0.0005 \\
    & Batch size & 128 \\
    & Standard Augmentations & Random crop, random horizontal flip \\
    & Training epochs & 50 \\
    & Initial learning rate & 0.1 \\
    & Learning rate schedule & Epochs per decay: 100, decay factor: 0.5  \\
\midrule
\multirow{8}{*}{SVHN}
    & Optimizer & SGD \\
    & Momentum & 0.9 \\
    & Weight-decay & 0.0005 \\
    & Batch size & 128 \\
    & Standard augmentations & None \\
    & Training epochs & 40 \\
    & Initial learning rate & 0.1 \\
    & Learning rate schedule & Epochs per decay: 100, decay factor: 0.5  \\
\midrule
\multirow{8}{*}{ImageNet-100}
    & Optimizer & SGD \\
    & Momentum & 0.9 \\
    & Weight-decay & 0.0005 \\
    & Batch size & 256 \\
    & Standard augmentations & Random crop, horizontal flip, and color jitter \\
    & Training epochs & 100 \\
    & Initial learning rate & 0.1 \\
    & Learning rate schedule & Epochs per decay: 100, decay factor: 0.5  \\
\bottomrule
\end{tabular}}
\end{table}
\begin{table}[!ht]
\centering\caption{%
    Default hyperparameter settings of defenses.
}\label{tab:config:defense}
\adjustbox{scale=0.8}{%
\begin{tabular}{lll}
\toprule
Methods & Hyperparameter & Settings \\
\midrule
\multirow{4}{*}{Adversarial training~\cite{madry2017towards}}
    & Perturbation & \(\ell_\infty = 8/255\) \\
    & Steps size & \(\ell_\infty = 2/255\) \\
    & PGD steps & 10 \\
    & Training epochs & 200 \\
\midrule
CutOut~\cite{devries2017improved} /
CutMix~\cite{yun2019cutmix} /
MixUp~\cite{zhang2017mixup}
    & Training epochs & 60 \\
\midrule
\multirow{3}{*}{Gaussian~\cite{liu2023image}}
    & Kernel size & 3 \\
    & Standard deviation & 0.1 \\
    & Training epochs & 100 \\
\midrule
\multirow{2}{*}{JPEG~\cite{liu2023image}}
    & Quality & 10 \\
    & Training epochs & 200 \\
\midrule
\multirow{2}{*}{BDR~\cite{liu2023image}}
    & Number of bits & 2 \\
    & Training epochs & 200 \\
\midrule
\multirow{3}{*}{UEraser-Lite~\cite{qin2023learning}}
    & PlasmaBrightness / PlasmaContrast & p = 0.5 \\
    & ChannelShuffle & p = 0.5 \\
    & Training epochs & 200 \\
\midrule
\multirow{4}{*}{UEraser-Max~\cite{qin2023learning}}
    & PlasmaBrightness / PlasmaContrast & p = 0.5 \\
    & ChannelShuffle & p = 0.5 \\
    & Number of Repeats \(K\) & 5 \\
    & Training epochs & 300 \\
\midrule
\multirow{4}{*}{AVATAR~\cite{dolatabadi2023devil}}
    & Diffusion sampler & Score-SDE \\
    & Starting step / Total diffusion steps & 60 / 1000 \\
    & Pre-trained model & \href{https://github.com/yang-song/score_sde_pytorch}{Official pretrained} \\
    & Training epochs & 200 \\
\bottomrule
\end{tabular}}
\end{table}

\section{Additional Results}

\begin{figure}[ht]
    \centering
    \newcommand{\curveplot}[2]{%
        \begin{subfigure}{0.245\linewidth}
            {\includegraphics[
                width=\linewidth, trim=0pt 10pt 40pt 30pt, clip
            ]{fig/pltacc/#1.pdf}}
            \caption{#2.}
        \end{subfigure}}
    \curveplot{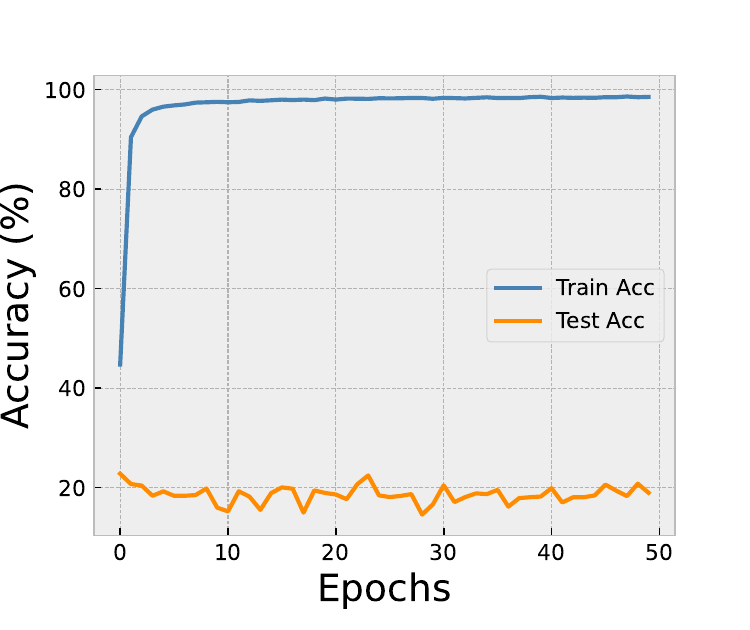}{EM without defense}
    \curveplot{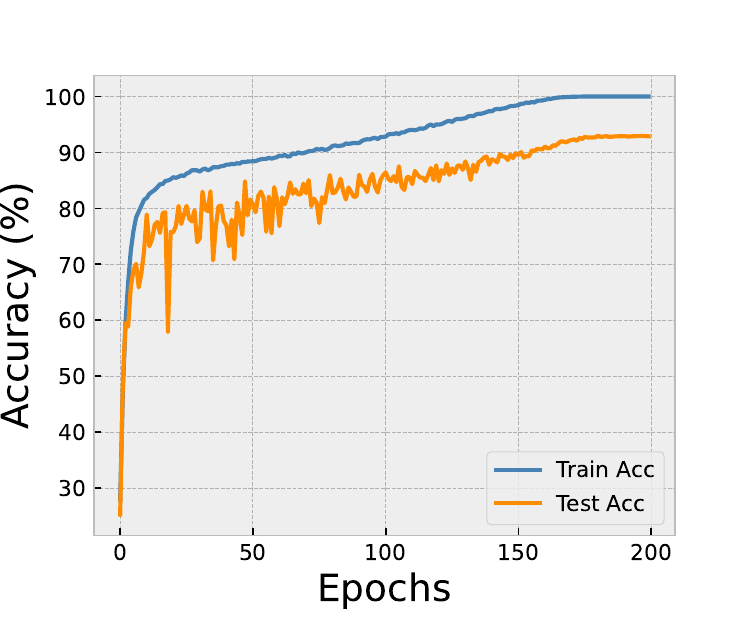}{EM with ISS}
    \curveplot{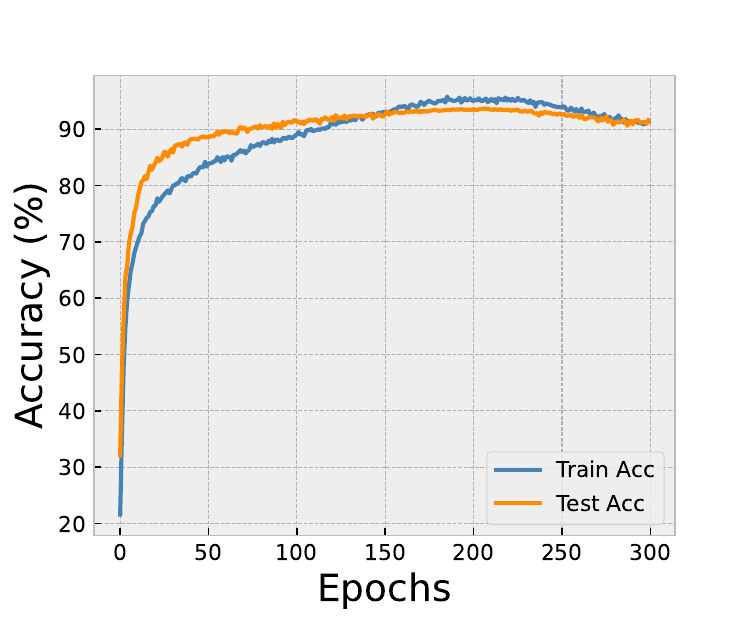}{EM with U-Max}
    \curveplot{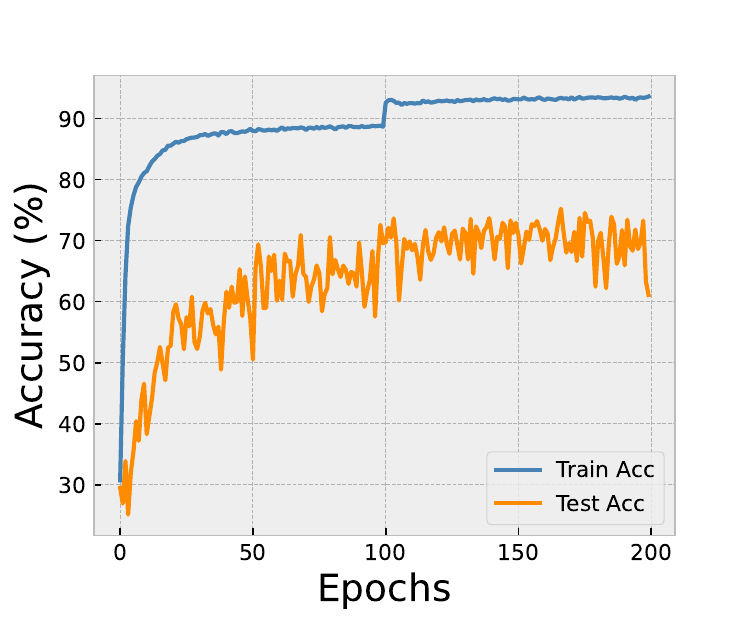}{EM with AVATAR}
    \curveplot{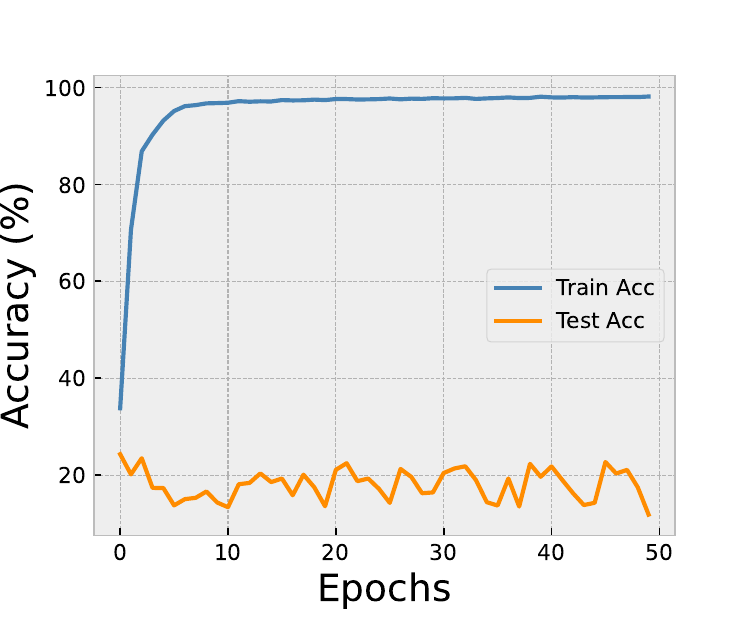}{REM without defense}
    \curveplot{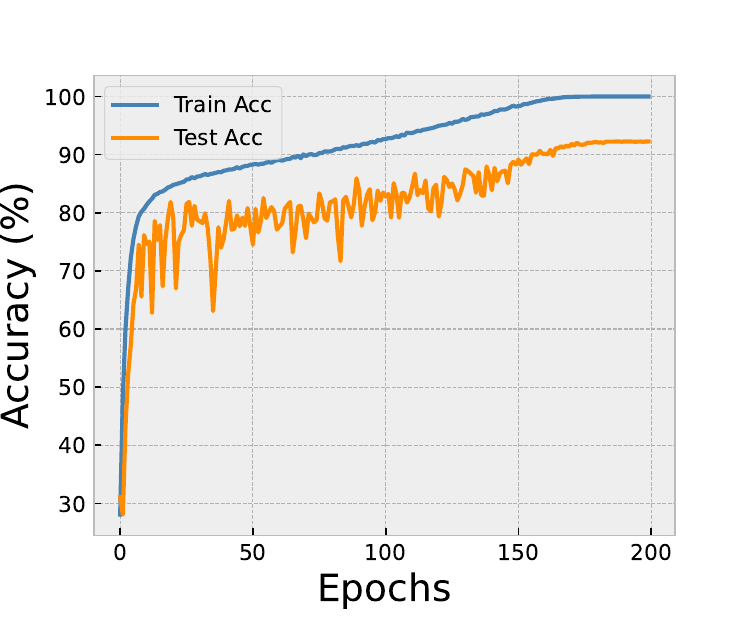}{REM with ISS}
    \curveplot{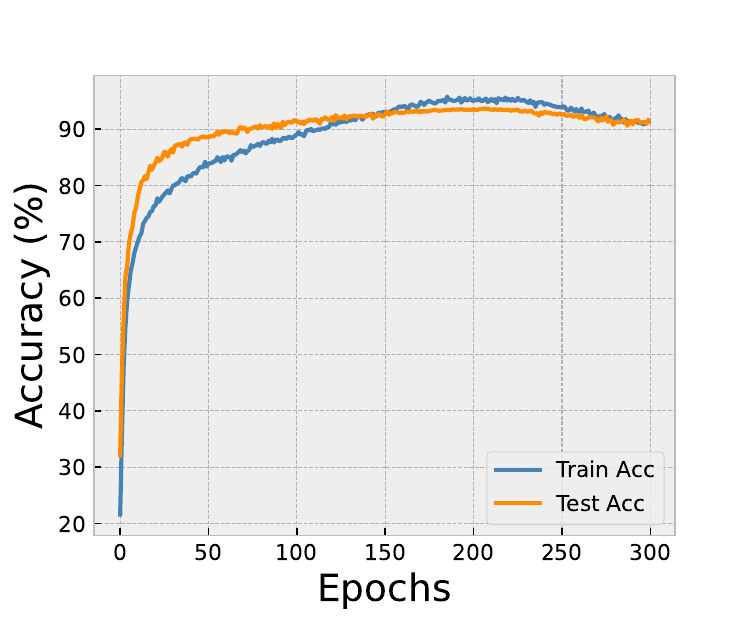}{REM with U-Max}
    \curveplot{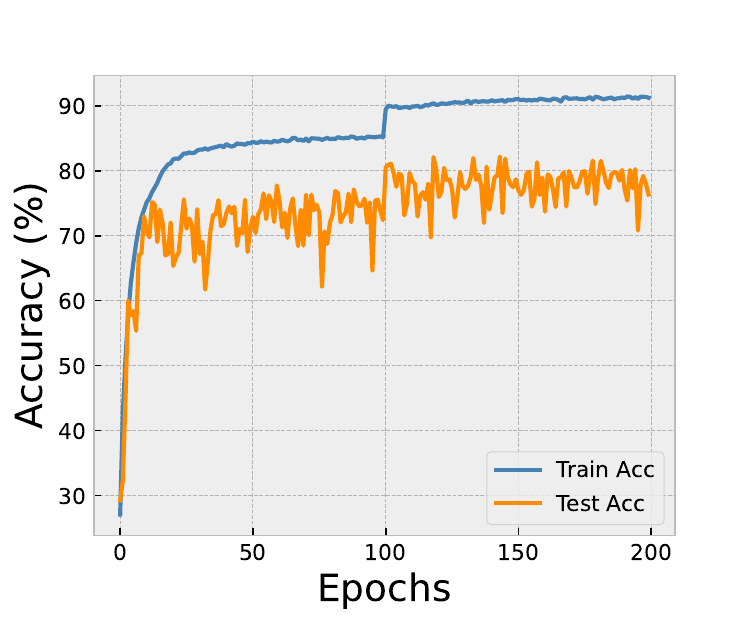}{REM with AVATAR}
    \curveplot{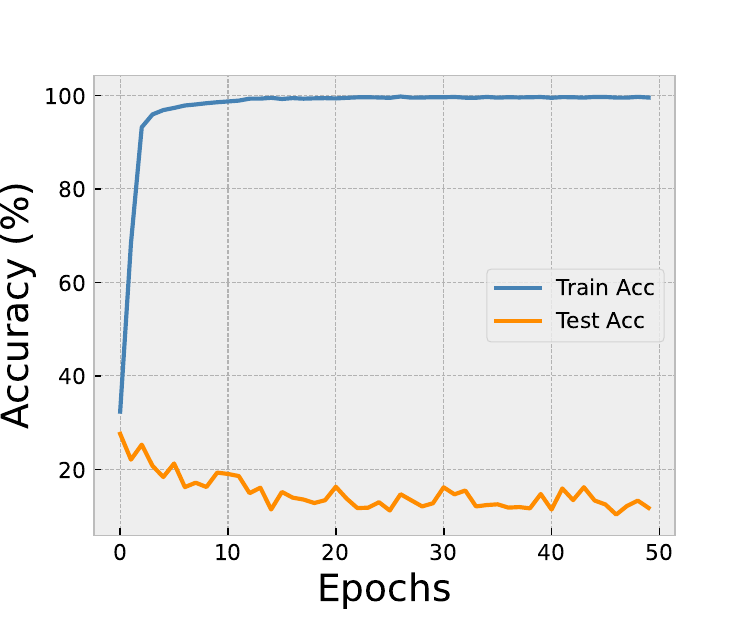}{LSP without defense}
    \curveplot{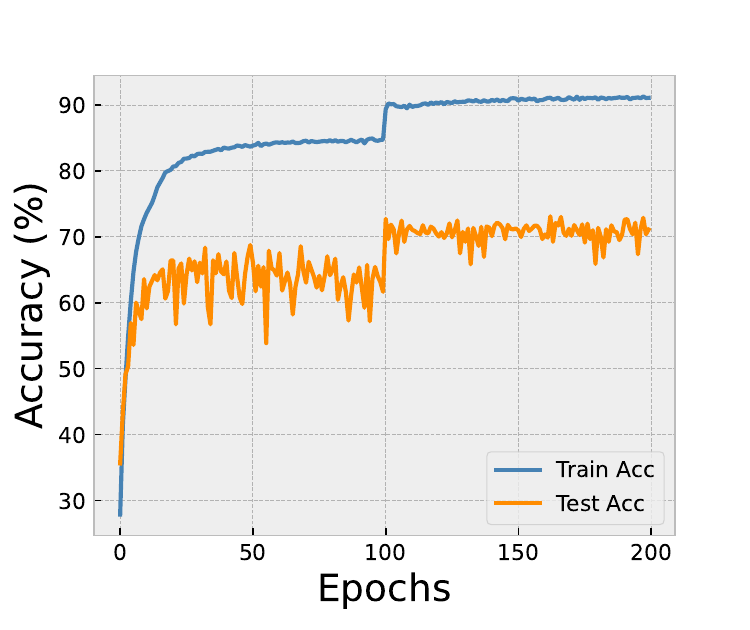}{LSP with ISS}
    \curveplot{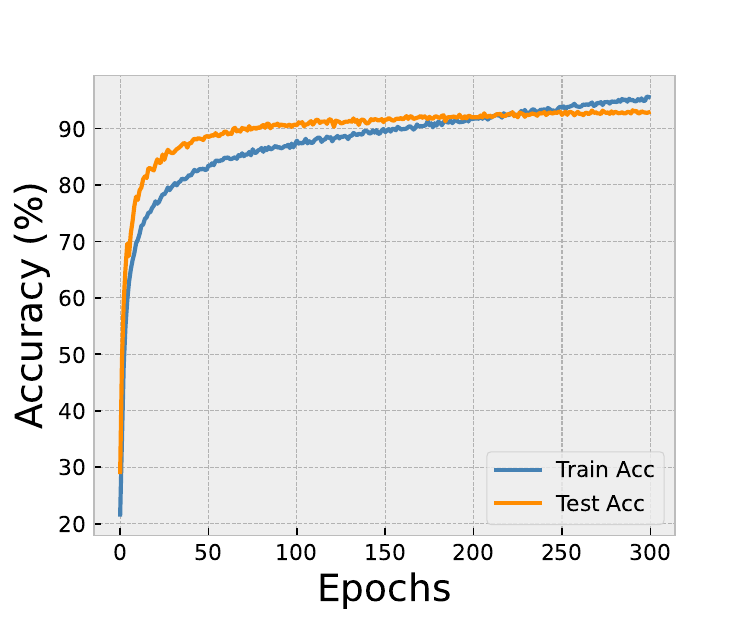}{LSP with U-Max}
    \curveplot{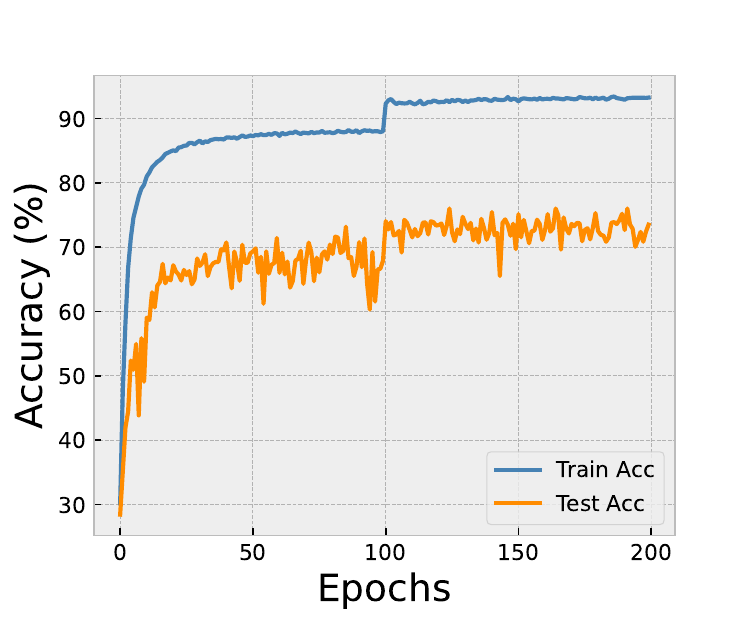}{LSP with AVATAR}
    \curveplot{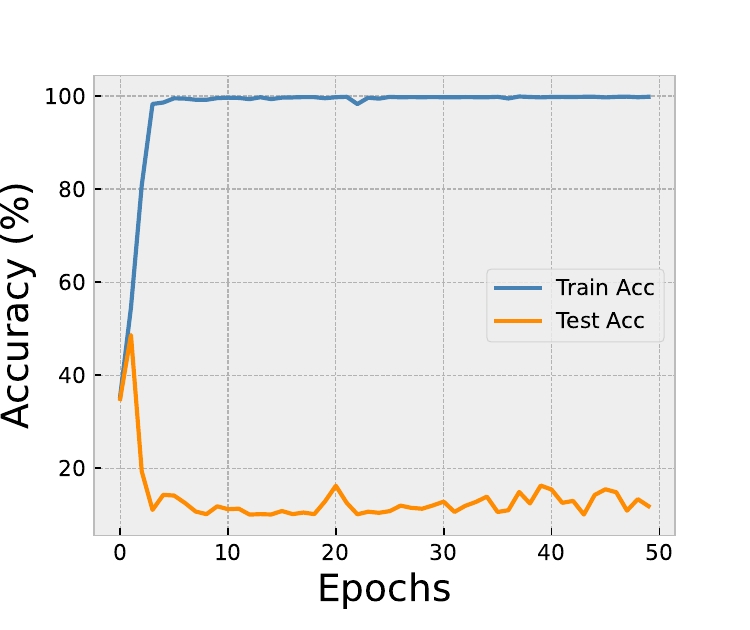}{AR without defense}
    \curveplot{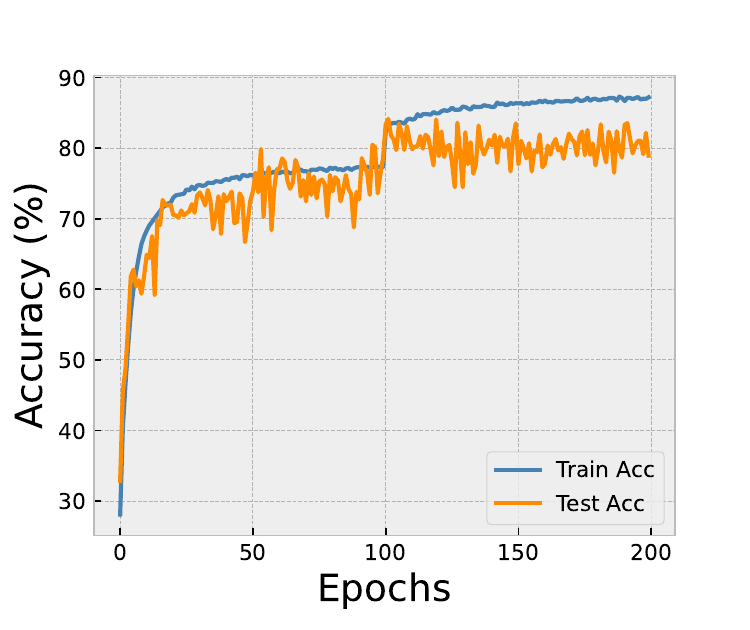}{AR with ISS}
    \curveplot{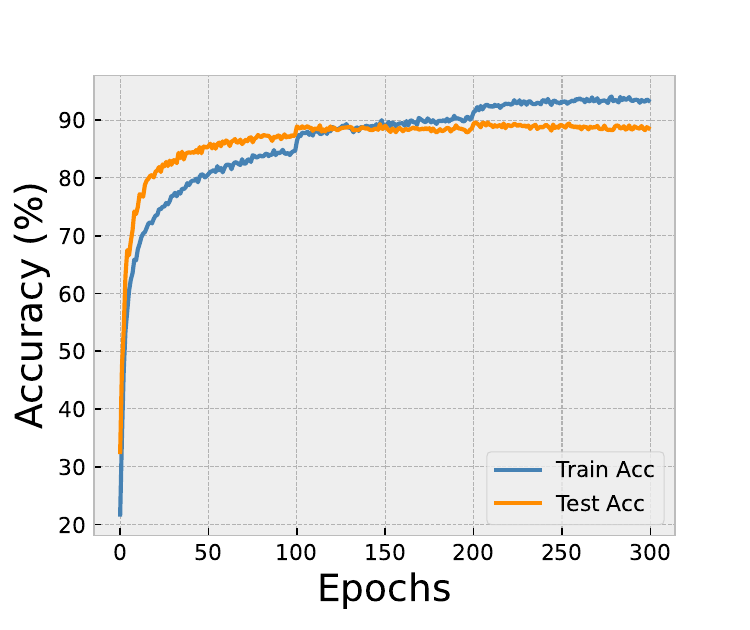}{AR with U-Max}
    \curveplot{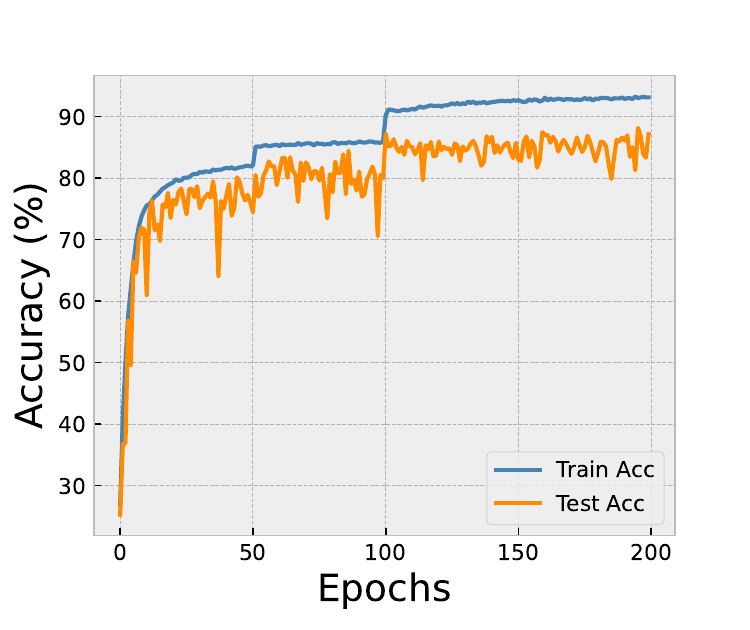}{AR with AVATAR}
    \caption{%
        Train and test accuracy curves
        during trainig for different defenses
        (standard training, ISS, UEraser-Max, and AVATAR)
        under different attacks
        (EM, REM, LSP, and AR)
        for CIFAR-10.
    }\label{fig:acc}
\end{figure}

\Cref{fig:acc}
shows the train and test accuracy curves
during training
with various defenses
under different attacks
for CIFAR-10.

\begin{figure}[ht]
    \includegraphics[width=0.90\linewidth]{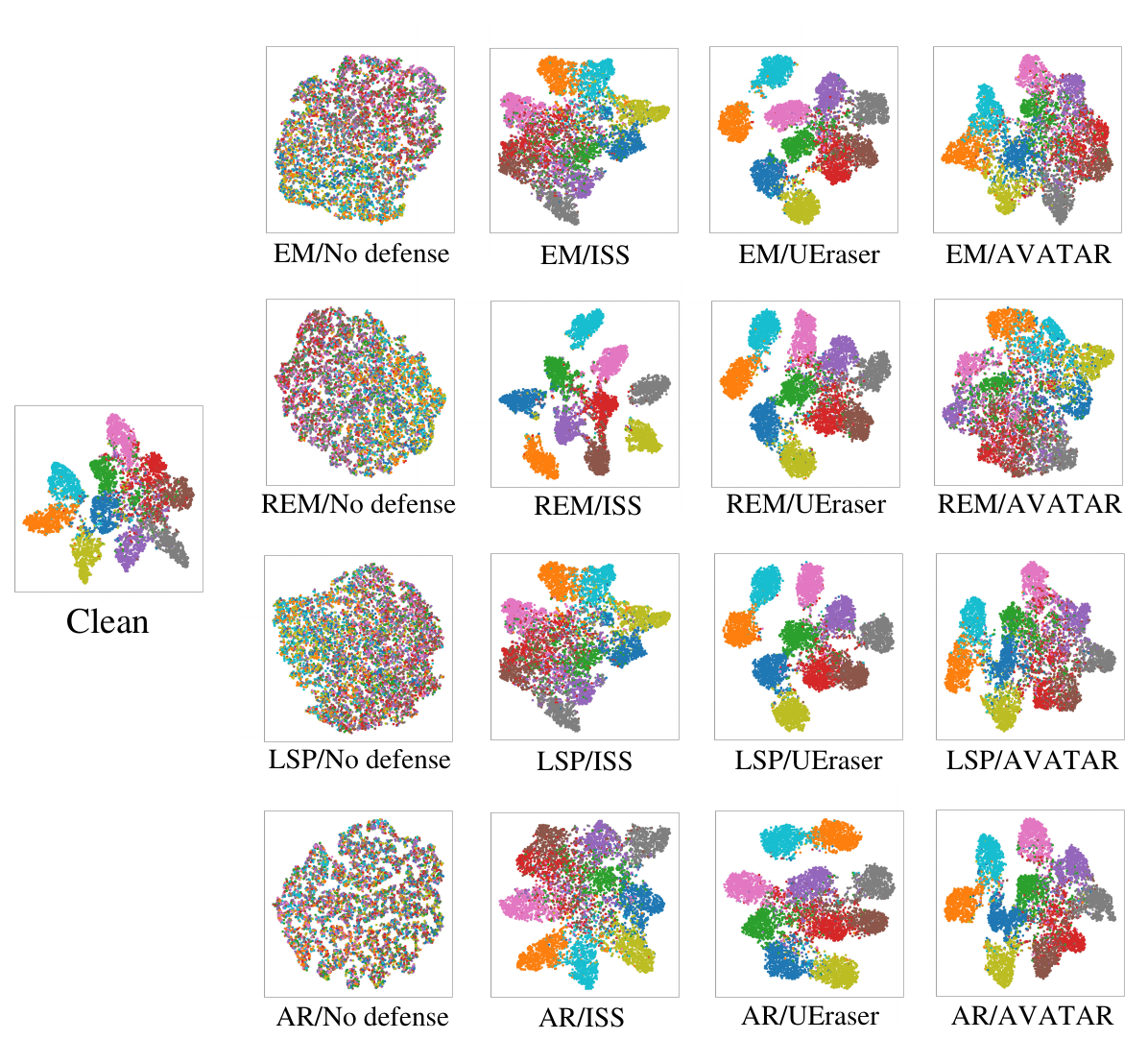}
    \caption{%
        The t-SNE visualization of the models'
        feature representations on the clean test set
        under additional attacks
        for CIFAR-10.
        ``U-Max'' denotes ``UEraser-Max''.
    }\label{fig:tsne:additional}
\end{figure}

\Cref{fig:tsne:additional}
shows the t-SNE visualization
of the models' feature representations
on the clean test set
for CIFAR-10.

\section{Limitations}\label{app:limitations}

APBench has mainly focused on
providing algorithms and evaluations
related to image data.
However,
such availability poisoning methods
may also be applicable to text, speech, or video domains.
In the future,
we plan to expand APBench
to include more domains,
aiming to establish a more comprehensive
and valuable benchmark
for personal privacy protection against deep learning.

\section{Ethics Statement}\label{app:ethics}

Similar to many other technologies,
the implementation of availability poisoning algorithms
can be used by users for both beneficial and malicious purposes.
We understand that these poisoning attack methods
were originally proposed to protect privacy,
but they can also be used
to generate maliciously data
to introduce model backdoors.
The benchmark aims to promote an understanding
of various availability poisoning attacks and defense methods,
as well as encourage the development of new algorithms in this field.
However,
we emphasize
that the use of these algorithms and evaluation results
should comply with ethical guidelines and legal regulations.
We encourage users to be aware
of the potential risks of the technology
and take appropriate measures
to ensure its beneficial use for both society and individuals.

\section{Computational Resources}\label{app:resources}

On NVIDIA V100 GPUs,
we used up to 0.3 / 1.0 / 3.2 / 4.7 GPU-hours per run
for standard training / ISS methods / UEraser-max / AT
on CIFAR-10, CIFAR-100 and SVHN.
As for ImageNet-100,
we used up to 1.0 / 1.6 / 4.9 / 5.8 GPU-hours
per run on NVIDIA A100
for standard training / ISS / UEraser-max / AT.
The preprocessing stage of AVATAR
took 4 GPU-hours on V100 on CIFAR-10.
For unsupervised learning,
the preprocessing cost of UEraser-lite and AVATAR
was approximately 0.4 / 4.0 GPU-hours on V100,
and we used up to 4.5 / 4.7 GPU-hours
per run for standard training / ISS on CIFAR-10.
All experiments in this paper
used approximately 3,800 GPU-hours on V100
and 1,100 GPU-hours on A100 in total.

\section{Licenses}\label{app:licenses}

APBench is open source,
and the source code is available at
\url{https://github.com/lafeat/apbench}.
\Cref{tab:licenses}
provides the licenses
of the derived implementations
of the original algorithms and datasets.
\begin{table}[ht]
\centering\caption{%
    Licenses of the datasets and codebases used in this paper.
}\label{tab:licenses}
\adjustbox{scale=0.8}{%
\begin{tabular}{lll}
    \toprule
    Name & License & URL \\
    \midrule
    PyTorch & BSD
        & \href{https://github.com/pytorch/pytorch}{GitHub: pytorch/pytorch} \\
    DC & \tna{}
        & \href{https://github.com/kingfengji/DeepConfuse}{GitHub: kingfengji/DeepConfuse} \\
    NTGA & Apache-2.0
        & \href{https://github.com/lionelmessi6410/ntga#unlearnable-datasets}
        {GitHub: lionelmessi6410/ntga} \\
    EM & MIT
        & \href{https://github.com/HanxunH/Unlearnable-Examples}{GitHub: HanxunH/Unlearnable-Examples} \\
    HYPO & MIT
        & \href{https://github.com/TLMichael/Delusive-Adversary}{GitHub: TLMichael/Delusive-Adversary} \\
    TAP & MIT
        & \href{https://github.com/lhfowl/adversarial_poisons}{GitHub: lhfowl/adversarial\_poisons} \\
    REM & MIT
        & \href{https://github.com/fshp971/robust-unlearnable-examples}
        {GitHub: fshp971/robust-unlearnable-examples} \\
    LSP & \tna{}
        & \href{https://github.com/dayu11/Availability-Attacks-Create-Shortcuts}
        {GitHub: dayu11/Availability-Attacks-Create-Shortcuts} \\
    AR & MIT
        & \href{https://github.com/psandovalsegura/autoregressive-poisoning}
        {GitHub: psandovalsegura/autoregressive-poisoning} \\
    OPS & Apache-2.0
        & \href{https://github.com/cychomatica/One-Pixel-Shotcut}
        {GitHub: cychomatica/One-Pixel-Shotcut} \\
    UCL & MIT
        & \href{https://github.com/kaiwenzha/contrastive-poisoning}{GitHub: kaiwenzha/contrastive-poisoning} \\
    TUE & \tna{}
        & \href{https://github.com/renjie3/TUE}{GitHub: renjie3/TUE} \\
    ISS & \tna{}
        & \href{https://github.com/liuzrcc/ImageShortcutSqueezing}{GitHub: liuzrcc/ImageShortcutSqueezing} \\
    DiffPure & NVIDIA
        & \href{https://github.com/NVlabs/DiffPure}{GitHub: NVlabs/DiffPure} \\
    CIFAR-10 & \tna{}
        & \url{https://www.cs.toronto.edu/~kriz/cifar.html} \\
    CIFAR-100 & \tna{}
        & \url{https://www.cs.toronto.edu/~kriz/cifar.html} \\
    SVHN & \tna{}
        & \url{http://ufldl.stanford.edu/housenumbers} \\
    ImageNet-100 & \tna{}
        & \href{https://github.com/TerryLoveMl/ImageNet-100-datasets}{GitHub: TerryLoveMl/ImageNet-100-datasets} \\
    \bottomrule
\end{tabular}}
\end{table}

\end{document}